\documentclass[lettersize,journal]{IEEEtran}
\usepackage{amsmath,amsfonts}
\usepackage{algorithmic}
\usepackage{array}
\usepackage[caption=false,font=normalsize,labelfont=sf,textfont=sf]{subfig}
\usepackage{textcomp}
\usepackage{stfloats}
\usepackage{url}
\usepackage{verbatim}
\usepackage{graphicx}
\usepackage{booktabs}
\def\BibTeX{{\rm B\kern-.05em{\sc i\kern-.025em b}\kern-.08em
    T\kern-.1667em\lower.7ex\hbox{E}\kern-.125emX}}
\usepackage{balance}
\begin{document}
\title{Scale-aware neural calibration for wide swath altimetry observations}
\author{Quentin Febvre, ~\IEEEmembership{Student,~IEEE,} Clément Ubelmann, Julien Le Sommer, Ronan Fablet%
\thanks{Thanks to IDRIS, Ifremer, CNES, CLS}}

\markboth{Transactions on geoscience and remote sensing}%
{Scale aware deep learning for wide swath altimetry calibration}

\maketitle

\begin{abstract}
	Sea surface height (SSH) is a key geophysical parameter for monitoring and studying meso-scale surface ocean dynamics. For several decades, the mapping of SSH products at regional and global scales has relied on nadir satellite altimeters, which provide one-dimensional-only along-track satellite observations of the SSH.  
	The Surface Water and Ocean Topography (SWOT) mission  deploys a new sensor that acquires for the first time wide-swath two-dimensional observations of the SSH. This provides new means to observe the ocean at previously unresolved spatial scales. A critical challenge for the exploiting of SWOT data is the separation of the SSH from other signals present in the observations. In this paper, we propose a novel learning-based approach for this SWOT calibration problem. It benefits from calibrated nadir altimetry products and a scale-space decomposition adapted to SWOT swath geometry and the structure of the different processes in play.
	In a supervised setting, our method reaches the state-of-the-art residual error of $\approx$ 1.4cm while proposing a correction on the entire spectral from 10km to 1000km and using weaker constraints on the modeled error signal.
\end{abstract}


\section{Introduction}

Nadir altimeter satellites provide invaluable direct measurements of the sea surface height (SSH) to monitor sea surface dynamics. 
They have played a key role in better understanding ocean circulation and improving climate monitoring. Altimeter-derived SSH data are also of key interest for offshore activities, marine pollution monitoring or maritime traffic routing among others. 

However due to the sparse and irregular sampling associated with nadir altimeter constellations, a wide range of ocean processes from the mesoscale to the submesoscale range remains unresolved, typically for horizontal scales below 150 kilometers and time scales below 10 days. 
The recently launched SWOT mission, with its Ka-band radar interferometer (KaRIN) sensor, provides for the first time higher-resolution and two-dimensional snapshots of the SSH. Once this data is adequately processed, it will likely strongly impact our ability to observe and study upper ocean dynamics \cite{Peral_Esteban-Fernandez_2018}. 

\begin{figure}[!t]
    \begin{center}
        \includegraphics[width=\linewidth]{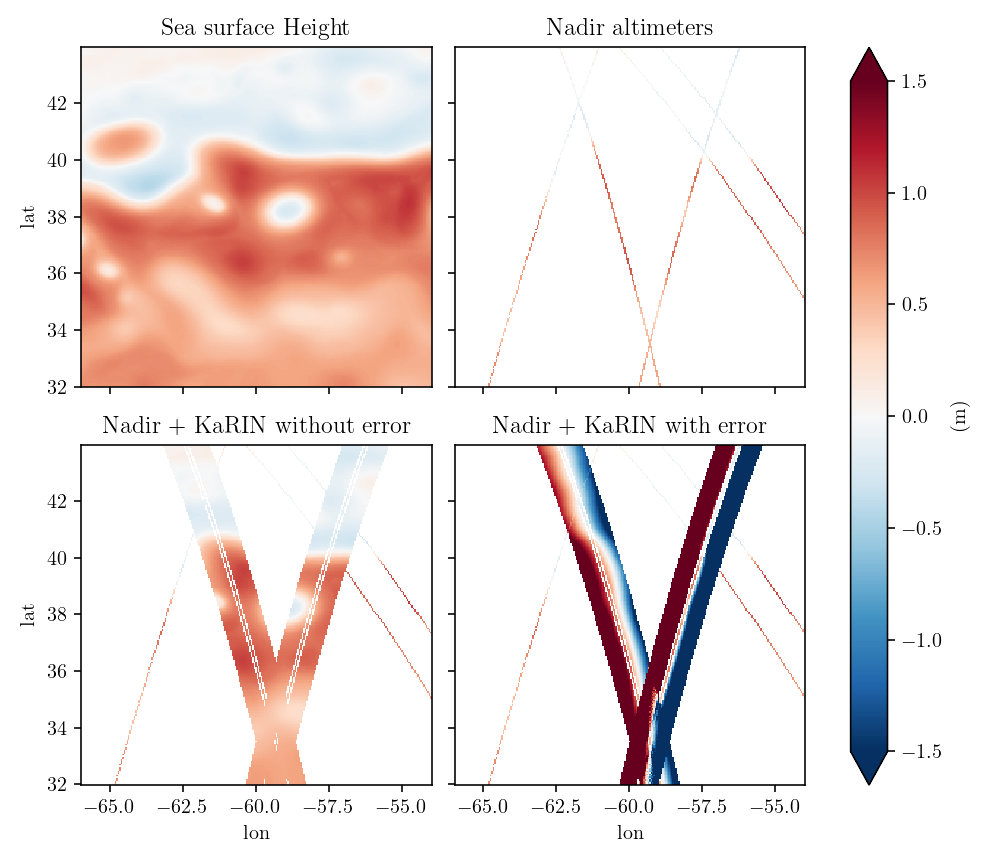}
    \end{center}
    \caption{\textbf{Observing System Simulation Experiment Cross-Calibration data:} \textit{Top left:} Sea surface height (SSH) on October 26th 2012 from NATL60 simulation dataset. \textit{Top right:} Calibrated NADIR pseudo-observations sampled using realistic orbits from the SSH are used to compute gridded product for the cross-calibration.\textit{Bottom-left:} NADIR + noise-free-KaRIN pseudo-observations, the  2{\sc d} sampled the SSH is the target of the cross-calibration.\textit{Bottom-right:} NADIR + noisy-KaRIN pseudo-observations, simulated errors added to the swath SSH constitute the uncalibrated input of the cross-calibration problem}
\label{fig:gridded}
\end{figure}

\begin{figure*}[!t]%
   \centering
    \subfloat[$f$]{{\includegraphics[width=.49\textwidth]{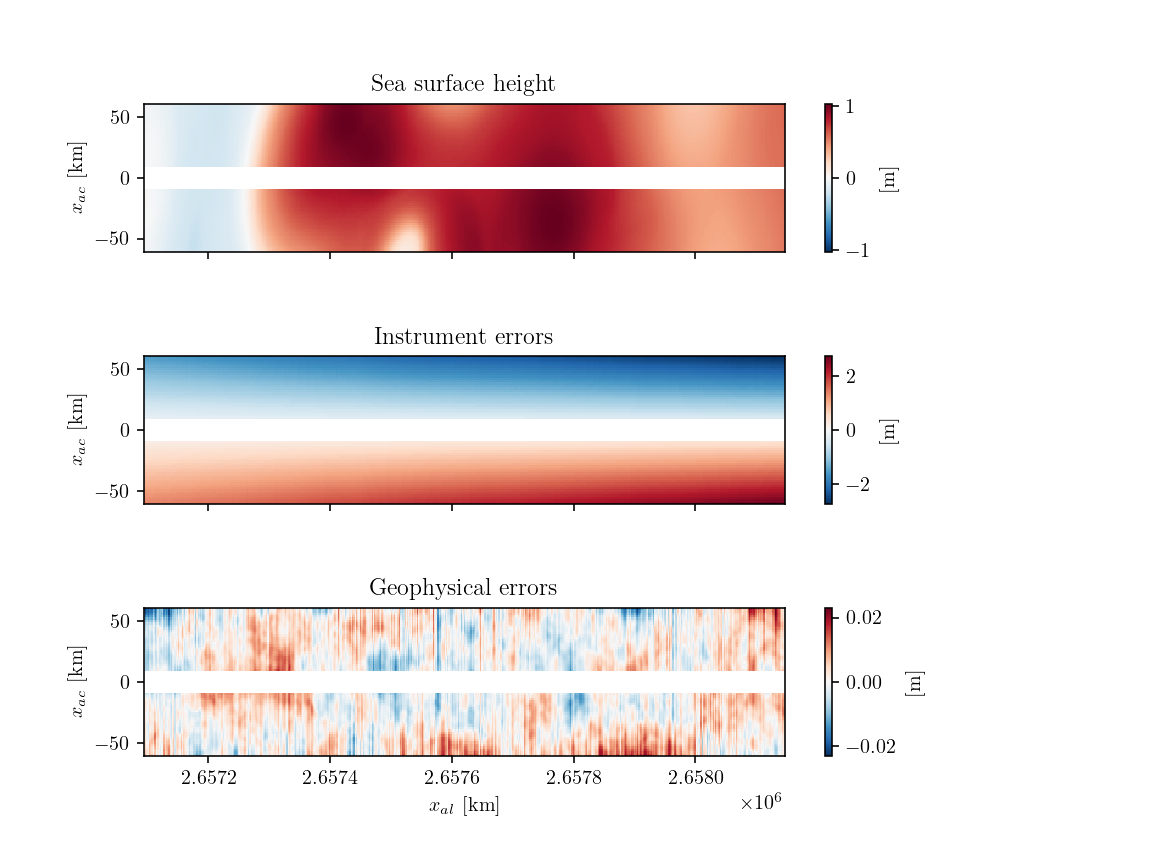} }}%
    \subfloat[$\mathcal{G}_{200km}((f)) - \mathcal{G}_{10km}(f)$]{{\includegraphics[width=.49\textwidth]{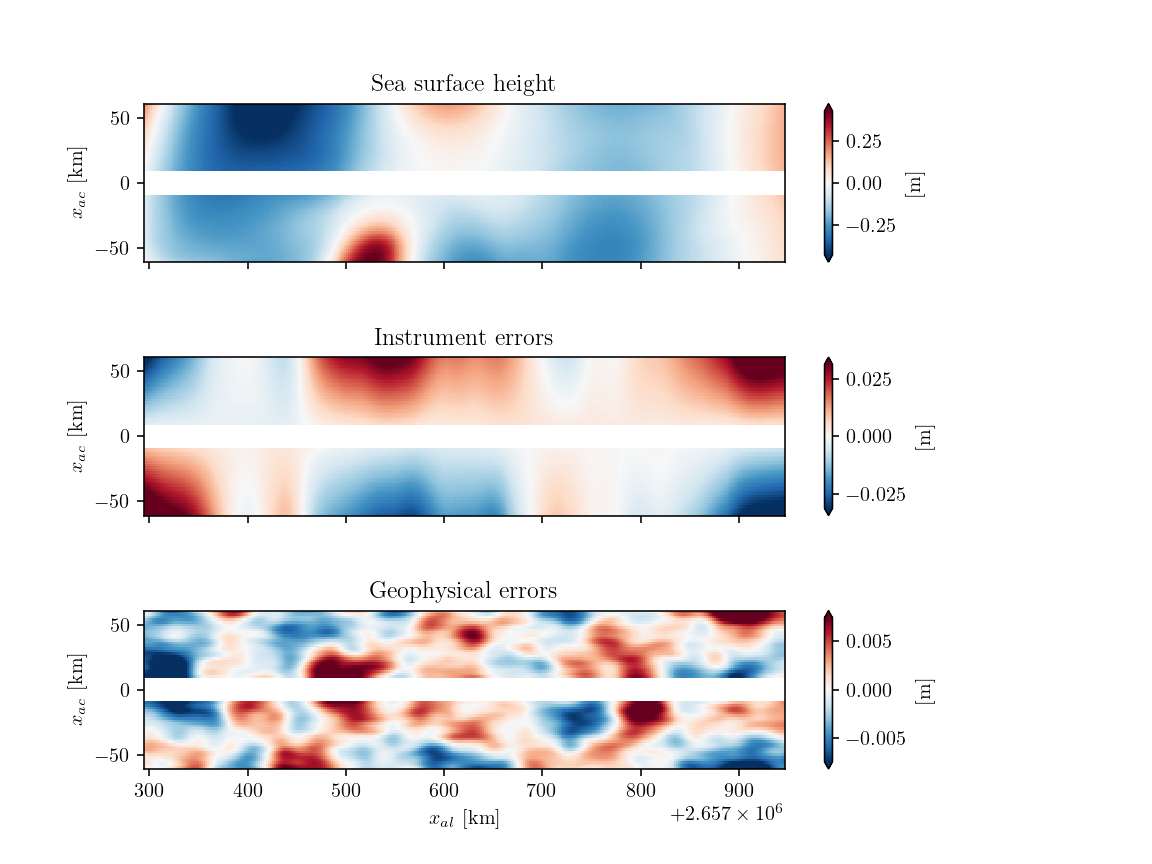} }}%
    \caption{\textbf{1000km segment of KaRIN observation components in swath geometry:}\textit{(a)} Looking at the three signals we see that the large scale instrument errors (middle) are predominant compared to the SSH (top) and geophysical error (bottom). \textit{(b)} Looking at the along-track scales between 10km and 200km, we see that the SSH is dominant w.r.t the error signals.}%
    \label{fig:filtered_swath_uncal_comp}%
\end{figure*}

As reported in Figure \ref{fig:gridded}, KaRIN data will be affected by instrument and geophysical errors  \cite{ubelmann_swot_nodate} and their exploitation requires to develop robust calibration schemes. We illustrate in Fig.\ref{fig:filtered_swath_uncal_comp} the two main error sources: instrument errors, especially roll errors, are expected to cause the dominant large-scale signal in both across-swath and along-swath directions; and geophysical errors, in particular due to wet-troposphere-induced delays\footnote{We refer the reader to Section \ref{subsec:altimetry} for the description of these error signals in raw KaRIN observations.}.  
The amplitude of these errors typically range from a few centimeters to a few meters in simulation, when the variability of the SSH for scales below 150km typically amounts to centimeters (See Fig. \ref{fig:gridded_impact}). This makes SWOT calibration a particularly challenging task in terms of signal-to-noise ratio. State-of-the calibration schemes \cite{Dibarboure_Ubelmann_Flamant_Briol_Peral_Bracher_Vergara_Faugere_Soulat_Picot_2022} rely on explicit spectral priors to address the calibration problem. 
The underlying hypotheses that one can linearly separate the SSH and the different error components may however impede the performance of such calibration methods. Here, we propose a novel learning-based approach. 
We leverage the computational efficiency of deep learning schemes with a scale-space decomposition \cite{Witkin_1984} adapted to the geometry of KaRIN observations. 

Our main contributions are as follows:
\begin{itemize}
\item{We state the cross-calibration of KaRIN altimetry observations as a learning problem using both raw KaRIN altimetry data and a gridded altimetry product as inputs to the neural network.}
\item{Our neural network architecture applies a scale-space decomposition scheme in the geometry of the KaRIN swath to improve the separability of the SSH and of the errors.}
\item{Numerical experiments using an Observing System Simulation Experiment (OSSE) demonstrate the relevance of the proposed approach and highlight the impact of the quality of the gridded altimetry product to retrieve finer-scale patterns in the calibrated KaRIN observations.}
\end{itemize}
This paper is organized as follows. Section \ref{sec:background} provides some background on related work. We introduce the considered data and case-study in Section \ref{sec:case_study}. 
Section \ref{sec:method} presents our method and we report numerical experiments in Section \ref{sec:results}. 
Section \ref{sec:conclusion} discusses further our main contributions.

\section{Background}
\label{sec:background}
\noindent
\subsection{Satellite altimeters}
In this paper, we address the cross-calibration of KaRIN observations, meaning that the proposed calibration scheme relies on external calibrated data. More specifically, we consider a constellation of nadir satellite altimeters of typically 4 to 7 satellites over the last decade. We recall that nadir altimeters provide can provide calibrated measurements of the SSH for medium to large scales along 1{\sc d} profiles corresponding to satellites' orbiting paths. 

By contrast, according to the mission's error budget specification \cite{Peral_Esteban-Fernandez_2018} the KaRIN instrument samples a two-dimensional swath of approximately 120km-wide with a 2km$\times$2km pixel resolution everywhere over the ocean.

In Figure \ref{fig:gridded}, we report simulated altimetry observations for both nadir altimeters and KaRIN along with the reference SSH issued from a numerical simulation dataset (see Section \ref{sec:case_study} for details). As an illustration of the complexity of calibration problem, the error signals completely occlude the SSH signal in the uncalibrated KaRIN observation.
Figure \ref{fig:filtered_swath_uncal_comp} illustrates further this point in the swath geometry. When focusing to along-track scales between 10km and 200km, the SSH signal becomes the main signal (Fig. \ref{fig:filtered_swath_uncal_comp}). This supports
both to consider a scale-space decomposition and to 
investigate a cross-calibration approach with
the exploitation of nadir-altimeter-derived altimetry products, which typically resolve horizontal scales longer than 80km.

\subsection{Interpolation of satellite-derived altimetry data}
\label{subsec:interpolation}

As mentioned previously, flying nadir altimeter constellations naturally advocate for considering the resulting interpolated SSH products as auxiliary data of interest to address the calibration of KaRIN observations. 

Regarding operational SSH products, we may distinguish the optimally-interpolated altimetry-derived product (DUACS) \cite{taburet_duacs_2019} and reanalysis products using ocean general circulation models to assimilate various observation datasets, including satellite altimetry and satellite-derived sea surface temperature data \cite{glorys_rea_2021}. Both types of products typically retrieve SSH dynamics on a global scale for horizontal scales above 150km and 10 days. 

Recently, a renewed interest has emerged in interpolation methods for ocean remote sensing data \cite{beauchamp_intercomparison_2020}\cite{fablet_end2end_2021}. Especially, deep learning schemes have emerged as appealing approaches to make the most of available observation datasets. Recent benchmarking experiments \cite{osse_data_challenge} point out significant potential gain compared with the above-mentioned operational products.

Here, we aim at investigating the extent to which the quality of L4 nadir-altimetry-derived SSH products may impact the calibration of KaRIN observations.

\subsection{Scale-space theory}
\label{subsec:scalespace}

Introduced in 1984 by A.P Witkin \cite{Witkin_1984}, 

The scale-space theory provides a mathematically-sound framework to decompose 2{\sc d} signals at different spatial scales \cite{Witkin_1984}. Gaussian scale-space methods are among the most widely used. They rely on applying Gaussian blur transformations with different standard deviations. This approach has been widely used in low-level image processing tasks  \cite{lindeberg1996edge,Lindeberg_2015}. 
Recent studies have used the scale-space theory in deep learning architectures \cite{Pintea_Tomen_Goes_Loog_van_Gemert_2021,Worrall_Welling_2019}. These neural networks better deal with multi-scale patterns in the data. 
Here, we draw inspiration from the scale-space framework to address the KaRIN calibration problem.
We design a scale-aware decomposition scheme as part of our learning approach with a view to 
better account for the different characteristic scales of the signals in play.

\subsection{Deep Learning for earth observation}
\label{subsec:dl}
\noindent
Convolutional neural networks are among the state-of-the-art neural architectures for image processing applications, including
image classification\cite{lecun98,resnet2016}, image in-painting\cite{liu_image_2018}, object detection\cite{yolo2016} and more.
They have also led to breakthroughs in remote sensing problems such as SAR image segmentation\cite{colin2021,colin_2022}, altimetry data interpolation \cite{fablet_joint_2021} and even sensor calibration \cite{li_convolutional_2022}.
The problem of multi-scale processing in neural networks has traditionally been tackled through the use of pooling layers in architectures such as the UNet \cite{Ronneberger_Fischer_Brox_2015}. As shown in the reported numerical experiments, these architectures do not reach a state-of-the-art performance for our KaRIN calibration problem. This advocates for the design of neural architectures better accounting for the key features of KaRIN observations.

\section{Data and Case-study}
\label{sec:case_study}
In this paper, we run an Observing System Simulation Experiment (OSSE), meaning that we rely on simulated data to apply and evaluate the proposed neural approach.
In this section, we present the different datasets considered in this OSSE.

\subsection{NATL60}
The simulation of the sea surface height field is taken from the NATL60 \cite{ajayi_spatial_2020} run of the NEMO ocean model. This simulation spans one year and covers the North Atlantic basin with a 1/60° spatial resolution. We more specifically use the data from a 12°$\times$12° domain over the Gulfstream ranging from the longitudes -66° to -52° and latitudes 32° to 44°.

\subsection{Nadir observations}
\label{subsec:altimetry}

In order to generate realistic nadir-altimeter pseudo-observations, we consider the real orbits of the years 2012 and 2013 of the four missions Topex-Poseidon, Jason 1, Geosat Follow-On, Envisat, as well as the 21-day cycle phase SWOT orbit from the SWOT simulator \cite{ubelmann_swot_nodate} project. The sampling of the nadir-altimeter pseudo-observations relies on the interpolation of the hourly SSH fields of the NATL60 run at the orbit coordinates. We consider nearest-neighbor interpolation in time and a bilinear interpolation in space.

\subsection{KaRIN observations}
The SWOT simulator also generates the swath coordinates on each side of the SWOT nadir. The swath spans from 10km to 60km off nadir with a 2km by 2km resolution. The SSH is then sampled on those coordinates the same way as the nadir observations.
Additionally, we also use the SWOT simulator in its "baseline" configuration to generate observation errors. 
Our simulation includes the systematic instrument errors with the roll, phase, timing and baseline dilation signals. Those signals have time-varying constant, linear or quadratic shape in the across track dimension. 
We also consider the geophysical error with the wet troposphere residual error as implemented in the simulator. 
We refer the reader to \cite{ubelmann_swot_nodate}
for a detailed presentation of the SWOT simulator.

\subsection{Gridded Altimetry Products}
\label{subsec:mapping}
\noindent
As explained in section \ref{subsec:interpolation}, we make use of interpolated SSH products based on
nadir altimetry data as inputs for our cross-calibration method.
We consider two interpolation schemes in our study:
\begin{itemize}
    \item the operational state-of-the-art based on optimal interpolation as implemented in the DUACS product \cite{taburet_duacs_2019}.
    \item a state-of-the-art neural interpolation scheme, referred to as 4DVarNet \cite{fablet_joint_2021}. This method is based on a trainable adaptation of the 4DVAR \cite{carrassi_data_2018} variational data assimilation method, and out-performs concurrent approaches in the considered OSSE setup \cite{osse_data_challenge}. We consider two 4DVarNet interpolation configurations, one using only nadir altimetry data \cite{Beauchamp_Febvre_Georgenthum_Fablet_2022}, one using jointly nadir altimetry and sea surface temperature data \cite{Fablet_Febvre_Chapron_2022}. We also include the latter as it significantly improves the reconstruction of the SSH at finer scales.
    
\end{itemize}

\section{Proposed Methodology}
\label{sec:method}
\noindent

This section presents the proposed methodology for the cross-calibration of raw KaRIN observations.
We design trainable neural architectures that take as inputs the uncalibrated KaRIN observations and the nadir-altimeter-derived gridded SSH products interpolated on the KaRIN swath. We train these architectures in a supervised manner on the reconstruction of the SSH on the KaRIN swath.
We first present an overview of the proposed neural architectures (Section \ref{subsec:neural_arch}). We then detail two specific components, namely the  scale-space decomposition block (Section \ref{subsec:scale_decomp}) and the swath-mixing layers (Section \ref{subsec:mixing}).

\subsection{Proposed neural architecture}
\label{subsec:neural_arch}
\noindent
\begin{figure*}
    \begin{center}
	    \includegraphics[width=\textwidth]{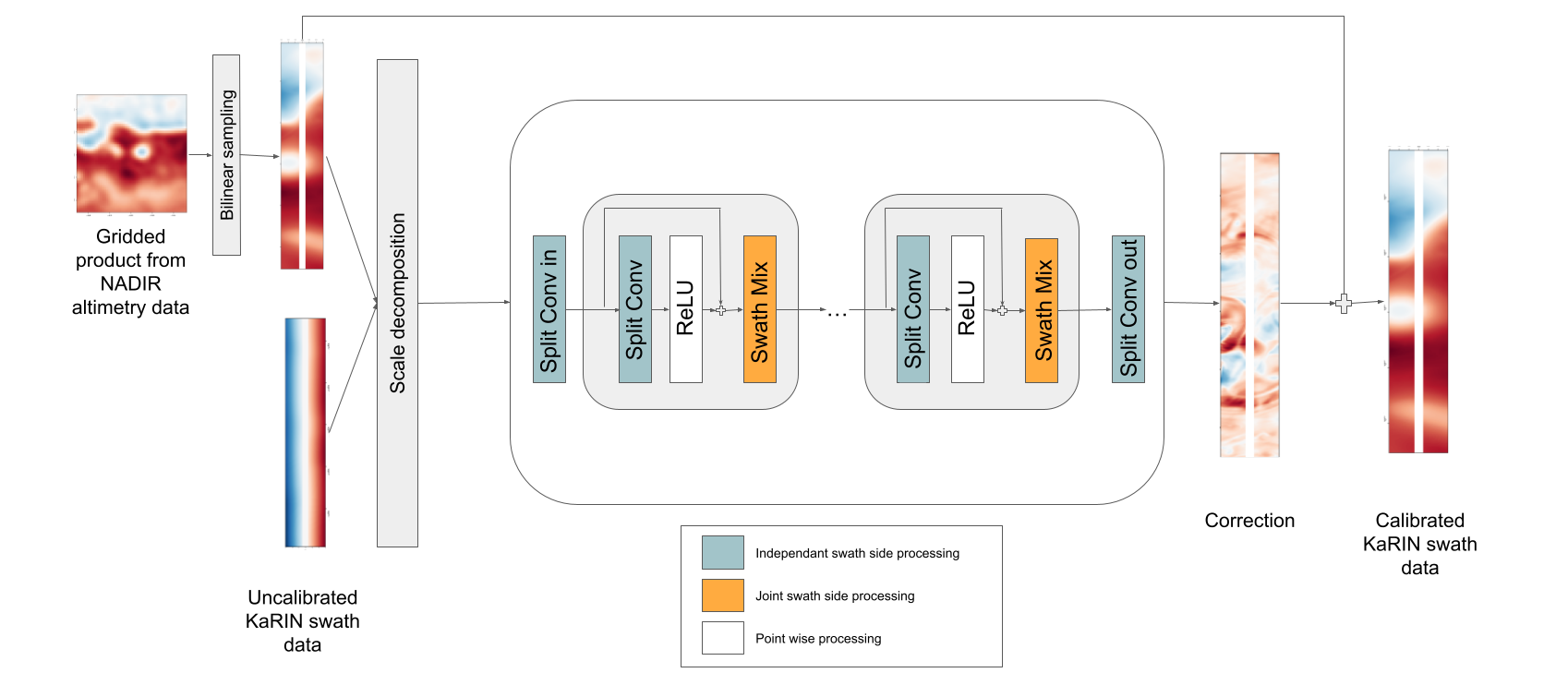}
    \end{center}
    \caption{\textbf{Overview of the proposed architecture:} From left to right: First, the nadir-based gridded product is interpolated on the swath segment. Then, it and the KaRIN observation  go through the scale decomposition scheme described in \ref{subsec:scale_decomp}. The scale components are stacked as channels and processed through the neural network. The green color of the "Split Conv" indicates that each side of the swath is processed independently by the convolution layer whereas the orange coloring of the "Swath Mix" layer tells that the whole data is processed jointly by the step (more details in \ref{subsec:mixing}). The final convolution computes a correction to be added to the gridded product for computing the calibrated KaRIN data}
    \label{fig:arch}	
\end{figure*}
The overall architecture considered is shown in figure \ref{fig:arch}. The scale-space decomposition block first decomposed independently the input L4 SSH products and KaRIN observations
into $N_s$-scale tensors, which we concatenate as the channel dimension.
This results into a tensor of shape $(2N_s, N_{al}, N_{ac})$ where $N_{al}$ and $N_{ac}$ are respectively the along track and across track sizes of the input swath section. The scale-space decomposition step is described in section \ref{subsec:scale_decomp}
A linear 2{\sc d} convolution layer follows. 
The data is then processed by a series of residual convolutional blocks composed of a convolution layer, a ReLU non-linearity \cite{Nair_Hinton_2019}, a skip connection and a swath-mixing layer as described in \ref{subsec:mixing}. A last linear convolution layer outputs a residual field, which we sum with the input gridded L4 SSH product to produce the calibrated KaRIN observation.

\subsection{Scale decomposition}
\label{subsec:scale_decomp}

\begin{figure}[!t]%
    \centering
    {\includegraphics[width=\linewidth]{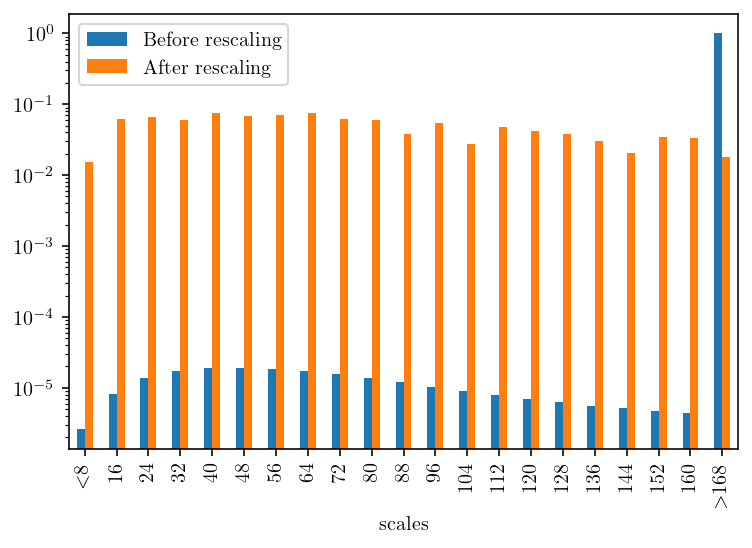} }%
    \caption{\textbf{Explained variance of scale components before and after re-scaling:} Each bar indicates how much each scale component of the uncalibrated KaRIN contributes to the total variance of the signal, we can see that before re-scaling (blue) there is four orders of magnitude between largest scale and the others. The learnt re-scaling allows for scale component to be spread within a single order of magnitude (orange), which is more suited to the downstream neural architecture.}%
    \label{fig:var_in_out}%
\end{figure}

We exploit a Gaussian scale-space to compute a scale-space decomposition of the fields provided as inputs to our neural architecture. For given scales $\sigma_1$ and $\sigma_2$, we extract the associated signal as the difference between filtered versions of the input signal using two Gaussian filters with standard deviation $\sigma_1$ and $\sigma_2$. We consider one-dimensional filters for the along-track direction. Formally, denoting 
 $\cal{G_{\sigma}}$ the 1-dimensional Gaussian blur operator with standard deviation $\sigma$ in the along track dimension, the considered scale-space decomposition of a signal $f$ given a sequence of increasing scales $[\sigma_1, \sigma_1, ..., \sigma_S]$ computes the following $S+1$ components: $[\mathcal{G}_{\sigma_1}(f), \mathcal{G}_{\sigma_2}(f) - \mathcal{G}_{\sigma_1}(f),...,\mathcal{G}_{\sigma_S}(f) - \mathcal{G}_{\sigma_{S-1}}(f), f - \mathcal{G}_{\sigma_S}]$
These different components are then considered as channels for the convolutionnal networks. In our experiments, we consider 20 scales in the along-track dimension evenly spaced from 8km to 160km. We discuss in section \ref{subsec:decomp_sens} how sensitive the proposed method is to the parameterization of the decomposition.
To account for scale-dependent energy levels in the computed scale-space representation (see fig. \ref{fig:var_in_out}), we introduce a batch normalization layer \cite{Ioffe_Szegedy_2015}. It re-scales each component to centered and unit-variance variables.
We illustrate in Fig.\ref{fig:var_in_out} the impact of the batch normalization step on the relative variance of the signal of each scale of the decomposition.

One may regard the proposed scale-scale decomposition as a convolutional block. Learning such a decomposition from data would however require very large convolutional filters, which does not seem 
realistic, or a deeper architecture with pooling layers that would require very efficient optimization given the quantity of data available. 

\subsection{Swath mixer block}
\label{subsec:mixing}

As observed in Figure \ref{fig:filtered_swath_uncal_comp}, the swath observed from the KaRIN sensor is not contiguous in the across-track dimension. The observation errors are however clearly correlated between the two sides of the swath. To exploit these correlations in our architecture, we design a swath-mixer block with two specific layers. 

To avoid convolution kernels to mix information from the two sides of the swath which could add some unwanted side effect, each side is processed separately by each convolution layer noted "Split Conv" in Fig. \ref{fig:arch}. Additionally, each convolution layer input is padded so that the height and width of the input remain unchanged throughout the network.

Besides, to combine relevant features from the two sides of the swath, we introduce a layer denoted as "Swath-mix" in Fig. \ref{fig:arch}. It implements a convolution layer after transposing the across-track dimension as a channel dimension. This idea of a mixer layer has been used in architectures such as the MLP-Mixer \cite{mlpmixer}, in which it has been shown to help with the expressiveness of neural networks.

We analyse in section \ref{subsec:ablation} the contribution of the mixing layer.

\section{Experimental results}
\label{sec:results}

\subsection{Setup}
\noindent
The results of this section have been computed using the one year ocean simulation NATL60, over the 12°x12° domain over the Gulfstream. The model evaluation is done on forty days in the inner 10°x10° region. The training of the mapping and calibration models are done on the remaining days.
The experimental setup used is the same as in \cite{osse_data_challenge}
The base configuration for our architecture uses three convolutional blocks with 128 channels as presented in Figure \ref{fig:arch}.
The supervised training loss is a weighted mean of the mean square errors for the reconstruction of the SSH, its gradient and its Laplacian.
The default scale-space decomposition used is made of twenty 8 kilometers band.
The calibration model is trained for 250 epochs with a annealing triangular cyclical learning rate \cite{Smith_2017}.

\subsection{Benchmarking experiments}
\label{subsec:main_res}
\noindent

\begin{table}[t]
\begin{center}
\begin{tabular}{lrr}
\toprule
 & RMSE (m) & RMSE $|| \nabla_{ssh} ||$ \\
\midrule
CalCNN & 1.39e-02 & 6.46e-03 \\
UNet & 2.34e-02 & 1.07e-02 \\
4DVarNet-5nad & 2.17e-02 & 9.57e-03 \\
\bottomrule
\end{tabular}

\end{center}
\caption{Residual error of the proposed calibration framework CalCNN}
\label{table:main}
\end{table}

We summarize our benchmarking experiments in Table \ref{table:main}.
We compare our approach, referred to as CalCNN, with a standard UNet \cite{Ronneberger_Fischer_Brox_2015} architecture. The latter uses as inputs the gridded altimetry product and the uncalibrated KaRIN observation stacked together as a 2{\sc d} field with 2 channels. We consider the same training configuration for this UNet model as for the CalCNN.
As baseline, we also consider the reconstruction performance for the KaRIN SSH issued from the 4DVarNet method using nadir-altimeter-only data. 
We evaluate all methods according to the following two metrics, the root mean squared error (RMSE) of the SSH field, and the RMSE of the amplitude of the gradients of the SSH field. 
Whereas the UNet fails to produce a better estimate than the nadir-only interpolation baseline, our CalCNN improves the estimation of the SSH and its gradient by over 35\% and brings the residual error below 1.4cm (Table \ref{table:main}).

\begin{figure}[!t]
    \centering
    \includegraphics[width=\linewidth]
    {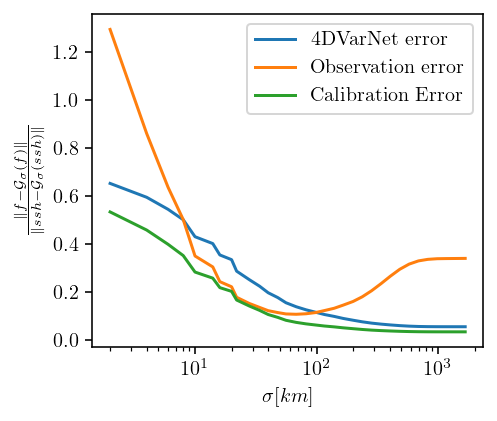}%
    \caption{{\bf Observation and reconstruction error for the SSH as a function of the spatial scales:}  The figure shows the error relative to the SSH at different scales for the inputs (Uncalibrated KaRIN in orange and nadir based interpolation in blue) and output (Calibrated KaRIN in green) of our method. The x axis correspond to the standard deviation of the Gaussian blur that was used to extract the low scale components. We can see the expected trend of the interpolation error of the nadir that is bigger at finer scale and becomes low w.r.t the SSH at large scale. Regarding the Uncalibrated KaRIN, we see that the error is lower than the interpolation only in the 10km-100km range. We see the Calibrated output of our method achieves lower error across all scales, successfully extracting relevant scale information from each signal.}
    \label{fig:err_scales}%
\end{figure}

In Figure \ref{fig:err_scales}, we further decompose the calibration error of the CalCNN w.r.t. the spatial scale using 1-dimensional Gaussian blurs as introduced in \ref{subsec:scale_decomp}. We draw a comparison with the 4DVarNet interpolation baseline and observation errors. The CalCNN reaches a lower error than both KaRIN observations and the interpolation baseline across all scales. At larger scales the error gets closer to the latter as instrument errors dominate the large-scale components of KaRIN observations. 
Interestingly, at scales lower than 10km, we still retrieve some improvement even though the observation error is quite high.
This can be explained by the fact that the high frequency errors on the KaRIN observations is easily separable from the underlying SSH signal.
Between 10-100km, our method successfully exploits the lower observation errors to improve the interpolation baseline.

\subsection{Ablation Study}
\label{subsec:ablation}
\noindent

\begin{table}[t]
\begin{center}
\begin{tabular}{lrr}
\toprule
 & RMSE (m) & RMSE $|| \nabla_{ssh} ||$ \\
xp &  &  \\
\midrule
CalCNN & 1.39e-02 & 6.46e-03 \\
CalCNN w/o skip connection & 2.17e-02 & 9.57e-03 \\
CalCNN w/o gridded product & 1.70e-01 & 2.47e-02 \\
CalCNN w/o scale decomposition & 2.17e-02 & 9.58e-03 \\
CalCNN w/o mixing layer & 1.94e-02 & 9.60e-03 \\
\bottomrule
\end{tabular}

\end{center}
\label{table:ablation}
\caption{Ablation results}
\end{table}

In this section we analyse further the contribution of the different components of our neural architecture. 
In Table \ref{table:ablation}, we report 
the performance metrics of the considered ablation study with the following models:
a model without skip connections, one without gridded product as input, one without the scale-space decomposition scheme (Sec. \ref{subsec:scale_decomp}) and one without the swath-mixer layers (Sec.\ref{subsec:mixing}. Overall, these four models lead to a significantly lower performance. 
The largest impact comes from the nadir-altimetry-only gridded product which provides large scale information about the SSH. And without which we lose an order of magnitude in the calibration errors.
Moreover, we can see that without the skip connections or scale decomposition, we fail to improve on the L4 gridded product.
Finally, we can note that we still get  ~10\% reduction of the RMSE w.r.t the L4 product without the mixing layer, however sharing the information between each side of the swath improves this gain three fold.

In Table \ref{table:size}, we show the sensitivity to the size of the network for the same training configuration. we compare the base architecture 3x128 (3 convolution blocks with 128 channels) with a linear operator, as well as a smaller network 1x32 and a bigger one 5x512. The linear version fails to extract geophysical information from the uncalibrated information. This further points out how challenging the considered calibration task is. Interestingly, our architecture leads to a similar performance for different complexity levels. The smaller and larger architectures leads to a slight increase in the residual error but the smaller model shows a slight improvement in the gradient reconstruction and spatial resolution.
Overall, these results support the robustness of the proposed learning-based approaches and the conclusions we raise in section \ref{subsec:main_res} are not very sensitive to the hyper-parameters of our network architecture.
\begin{table}[t]
\begin{center}
\begin{tabular}{lrr}
\toprule
 & RMSE (m) & RMSE $|| \nabla_{ssh} ||$ \\
xp &  &  \\
\midrule
128x3 (Ref) & 1.39e-02 & 6.46e-03 \\
Linear & 2.13e-02 & 1.02e-02 \\
32x1 & 1.44e-02 & 6.22e-03 \\
512x5 & 1.49e-02 & 7.19e-03 \\
\bottomrule
\end{tabular}

\end{center}
\caption{Impact of network size}
\label{table:size}
\end{table}

\subsection{Gridded product sensitivity} 
\label{subsec:gridded_sens}
\noindent

\begin{figure}
    \begin{center}
        \includegraphics{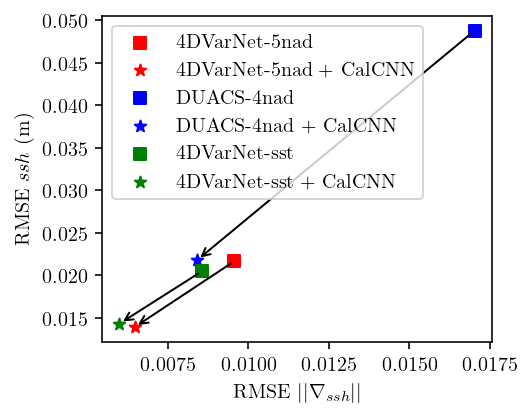}
    \end{center}
    \caption{{\bf Impact of the nadir-based gridded product on the CalCNN output:} The figure shows the RMSE and the RMSE of the $|| \nabla_{ssh} ||$ of the calibrated observation (stars) and their associated nadir-based gridded products (squares). We see an improvement brought by the CalCNN of both the RMSE and the RMSE of the $|| \nabla_{ssh} ||$ in the same direction for all three gridded products. This improvement can be interpreted as the relevant information extracted from the uncalibrated KaRIN by the CalCNN. The biggest relative improvement for DUACS (blue) which doesn't uses the SWOT's nadir altimeter, and the smallest one for the 4DVarNet that uses the SST.}
    \label{fig:gridded_impact}
\end{figure}

We analyze further how the quality of nadir-altimetry-only gridded product affects the calibration performance.
In Figure \ref{fig:gridded_impact}, we display the improvement in RMSE of the SSH on the swath and of the gradients of the SSH obtained by our CalCNN for the three gridded products introduced in Sec. \ref{subsec:interpolation}.

For all three interpolated products, the proposed calibration method improves the reconstruction of the SSH for the KaRIN swath from the joint analysis of the interpolation product and raw KaRIN observations. We report the larger improvement for DUACS product. This relates to the spectral overlap between the SSH information of the uncalibrated KaRIN and SWOT's NADIR. The associated calibration performance remains however significantly worse than that of the two 4DVarNet products, which may relate to the worse interpolation performance of DUACS product \cite{fablet_end2end_2021,osse_data_challenge}. 
When comparing the impact of the two 4DVarNet products, the results are more nuanced. The 4DVarNet-SST product leads to better metrics. The difference of RMSE is greatly reduced after calibration whereas the gap in RMSE of the gradients is conserved.
This could be interpreted as the gain of RMSE we get from using the SST can be obtained from the uncalibrated KaRIN, however some of the gradients we reconstruct through the SST are not easily extracted from the observations.
Overall this shows interesting relations between the redundant information in the uncalibrated KaRIN and the interpolated products.

\subsection{Sensitivity to the scale-space decomposition}
\label{subsec:decomp_sens}
\noindent
In Table \ref{table:scale_dec}, we display the calibration metrics for different scale-space decompositions. We vary the number of scales considered and the spacing between two consecutive scales. When considering the same scale range from 8km to 160km, we retrieve the best performance with 20 scales. But, even with only 5 scales evenly separated by 32 km, the performance decreases only by 3\%.
By contrast, when considering a scale separation of 8km but varying the number of scales, we note a more significant drop of performance (about 10\% in the residual RMSE). This suggests a greater sensitivity to the span of the scale-space decomposition than to the number and spacing of the components. 
However we still achieve less than 1.6cm residual error for any of the considered variations which is still a competitive calibration outcome.

\begin{table}[!t]
\begin{center}
	\begin{tabular}{llrr}
\toprule
 &  & RMSE (m) & RMSE $|| \nabla_{ssh} ||$ \\
$N_{band}$ & $\delta_{band}$ &  &  \\
\midrule
20 & 8 & 1.39e-02 & 6.46e-03 \\
40 & 4 & 1.44e-02 & 6.64e-03 \\
10 & 16 & 1.48e-02 & 6.75e-03 \\
5 & 32 & 1.41e-02 & 6.65e-03 \\
10 & 8 & 1.56e-02 & 6.81e-03 \\
40 & 8 & 1.54e-02 & 6.88e-03 \\
\bottomrule
\end{tabular}

\end{center}
\caption{Calibration metrics in function of the scale decomposition}
\label{table:scale_dec}
\end{table}

\section{Conclusion}
\label{sec:conclusion}
\noindent
We have proposed in this work a neural calibration approach which combines a scale-space decomposition of KaRIN observations and a convolutional architecture. This approach proves to be robust with a 
residual error below 1.5cm which can be compared with the 2cm residual error of the expected operational approaches performance although demonstrated globally using a different ocean simulation \cite{Dibarboure_Ubelmann_Flamant_Briol_Peral_Bracher_Vergara_Faugere_Soulat_Picot_2022}. While we can reach a satisfactory calibration performance using the operational nadir altimetry mapping product, our experiments highlight the potential benefit of ongoing effort on neural SSH interpolation schemes to further improve the retrieval of finer-scale features from KaRIN observations.
This naturally advocates for future work exploring jointly calibration and mapping problems for nadir and wide-swath altimeters, possibly combining our deep learning approach and variational mapping formulations introduced in \cite{Febvre_Fablet_Sommer_Ubelmann_2022}.
Finally, even though the calibration results shown here are promising, the generalization to real signals of our calibration operator trained on simulated data are yet to be demonstrated.

\bibliographystyle{IEEEbib}
\bibliography{biblio}


\end{document}